\documentclass[runningheads]{llncs}
\usepackage[T1]{fontenc}
\usepackage{graphicx,verbatim}
\usepackage{amsmath}
\usepackage[normalem]{ulem}
\useunder{\uline}{\ul}{}
\usepackage{array} 

\begin{document}
\title{CREATE-FFPE: Cross-Resolution Compensated and Multi-Frequency Enhanced FS-to-FFPE Stain Transfer for Intraoperative IHC Images}


%

\author{
    Yiyang Lin\textsuperscript{1,*}, 
    Danling Jiang\textsuperscript{2,*}, 
    Xinyu Liu\textsuperscript{1,*}, 
    Yun Miao\textsuperscript{3}, 
    and Yixuan Yuan\textsuperscript{1,†}
}  
\authorrunning{Yiyang Lin, Danling Jiang, Xinyu Liu, Yun Miao, and Yixuan Yuan}
\institute{
    \textsuperscript{1}Department of Electronic Engineering, The Chinese University of Hong Kong \\ 
    \email{yylin247@163.com, yxyuan@ee.cuhk.edu.hk} \\
    \textsuperscript{2}Department of Gastroenterology, Peking University Shenzhen Hospital \\ 
    \textsuperscript{3}Theory Lab, Central Research Institute, 2012 Labs, Huawei Technologies Co., Ltd. \\
    \email{miaoyun2@huawei.com}
}
\renewcommand{\thefootnote}{}
\footnotetext{\textsuperscript{*}These authors contributed equally.}
\footnotetext{\textsuperscript{†}Corresponding author: Yixuan Yuan.}

\maketitle              
\begin{abstract}
In the immunohistochemical (IHC) analysis during surgery, frozen-section (FS) images are used to determine the benignity or malignancy of the tumor. However, FS image faces problems such as image contamination and poor nuclear detail, which may disturb the pathologist's diagnosis. In contrast, formalin-fixed and paraffin-embedded (FFPE) image has a higher staining quality, but it requires quite a long time to prepare and thus is not feasible during surgery. To help pathologists observe IHC images with high quality in surgery, this paper proposes a \textbf{C}ross-\textbf{RE}solution compens\textbf{AT}ed and multi-frequency \textbf{E}nhanced FS-to-FFPE (CREATE-FFPE) stain transfer framework, which is the first FS-to-FFPE method for the intraoperative IHC images. To solve the slide contamination and poor nuclear detail mentioned above, we propose the cross-resolution compensation module (CRCM) and the wavelet detail guidance module (WDGM). Specifically, CRCM compensates for information loss due to contamination by providing more tissue information across multiple resolutions, while WDGM produces the desirable details in a wavelet way, and the details can be used to guide the stain transfer to be more precise. Experiments show our method can beat all the competing methods on our dataset. In addition, the FID has decreased by 44.4\%, and KID×100 has decreased by 71.2\% by adding the proposed CRCM and WDGM in ablation studies, and the performance of a downstream microsatellite instability prediction task with public dataset can be greatly improved by performing our FS-to-FFPE stain transfer. 

\end{abstract}
\section{Introduction}
\begin{figure}[t]
	\centerline{\includegraphics[width=1.0\columnwidth]{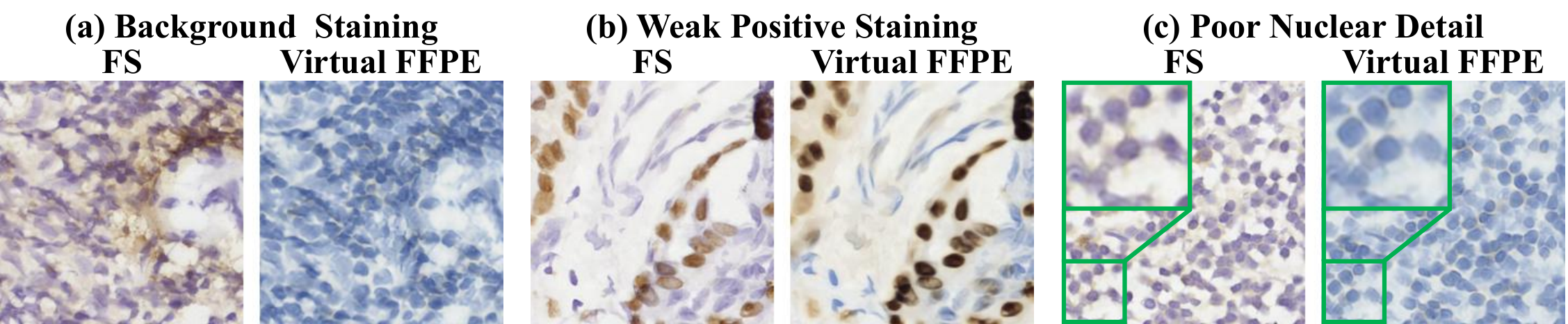}}
	\caption{\textbf{Examples of defects in FS images and our virtual FFPE.} The defects mainly include background staining, weak positive staining, and poor nuclear detail.
    }
	\label{intro}
\end{figure}

Rapid immunohistochemistry (IHC) staining using the frozen section (FS) is crucial in cancer surgery, which can facilitate the pathologist to determine whether the tumor is benign or malignant and whether the malignant tissue has been completely removed \cite{vahini2017intraoperative,chao2004use}. However, the FS slide faces problems such as background staining, weak positive staining, and poor nuclear detail, which may disturb the pathologists' diagnosis \cite{yigzaw2024review}. Specifically, in Fig. \ref{intro}, the brown background staining in (a) causes pathologists to misjudge negative nuclei as positive, the weak positive staining in (b) makes positive nuclei easily recognized as negative, and the poor nuclear detail in (c), especially the green zoom-in area, makes it difficult to observe details such as nuclear boundaries and nucleoli. In contrast, formalin-fixed and paraffin-embedded (FFPE) slides have higher quality and are capable of overcoming most of the problems in FS slides. However, the FFPE slide usually needs one to three days to prepare \cite{obaid2024comparison}. Given the time constraints of surgery, FFPE staining cannot be conducted during surgery. Therefore, transferring FS images to FFPE ones can allow pathologists to observe the IHC images with high quality, greatly reducing the misdiagnosis rate. 


Through generative adversarial networks (GAN) \cite{goodfellow2020generative} based method can transfer FS images to FFPE ones, they face two challenges. Firstly, the staining status (positive or negative, same as below) of the poorly stained FS image is often incorrectly distinguished by the network, leading to the false staining status of the transfer result. Since the staining status of the poorly stained image is usually consistent with its surrounding images, providing the well-stained surrounding images will help to better judge their staining status. Secondly, fine details such as nucleus boundaries and nucleoli are often poorly generated, sometimes being vague and sometimes missing. It is reported that most of these details are high-frequency information. Therefore, enhancing the processing and generation of high-frequency information can largely solve this problem.

Thus, we represent a \textbf{C}ross-\textbf{RE}solution compens\textbf{AT}ed and multi-frequency \textbf{E}nhanced FS-to-FFPE (CREATE-FFPE) stain transfer framework, which is the first FS-to-FFPE method on intraoperative IHC images. In CREATE-FFPE, to address the false generation of staining status in poorly stained FS images, we design a cross-resolution compensation module (CRCM) to additionally provide more well-stained tissue information across multiple resolutions surrounding the input image. Thus, the transfer result can maintain the staining status of the input well, thereby improving the accuracy of CREATE-FFPE. Moreover, to enhance the high-frequency generation, we propose a wavelet detail guidance module (WDGM). This module processes low-frequency and high-frequency information separately, preventing them from disturbing each other. Thus, the high-frequency details in WDGM can be generated well and used to guide CREATE-FFPE in achieving desirable details. As a result, in Fig. \ref{intro}, the virtual FFPE images correct almost all the defects of FS, markedly improving the FS image quality. Furthermore, when the stain image is scarce, CREATE-FFPE outperforms large-scale models with less computational cost and higher speed. 

\section{Method}
\subsection{The Overview of CREATE-FFPE}
\begin{figure}[t]
	\centerline{\includegraphics[width=1.0\columnwidth]{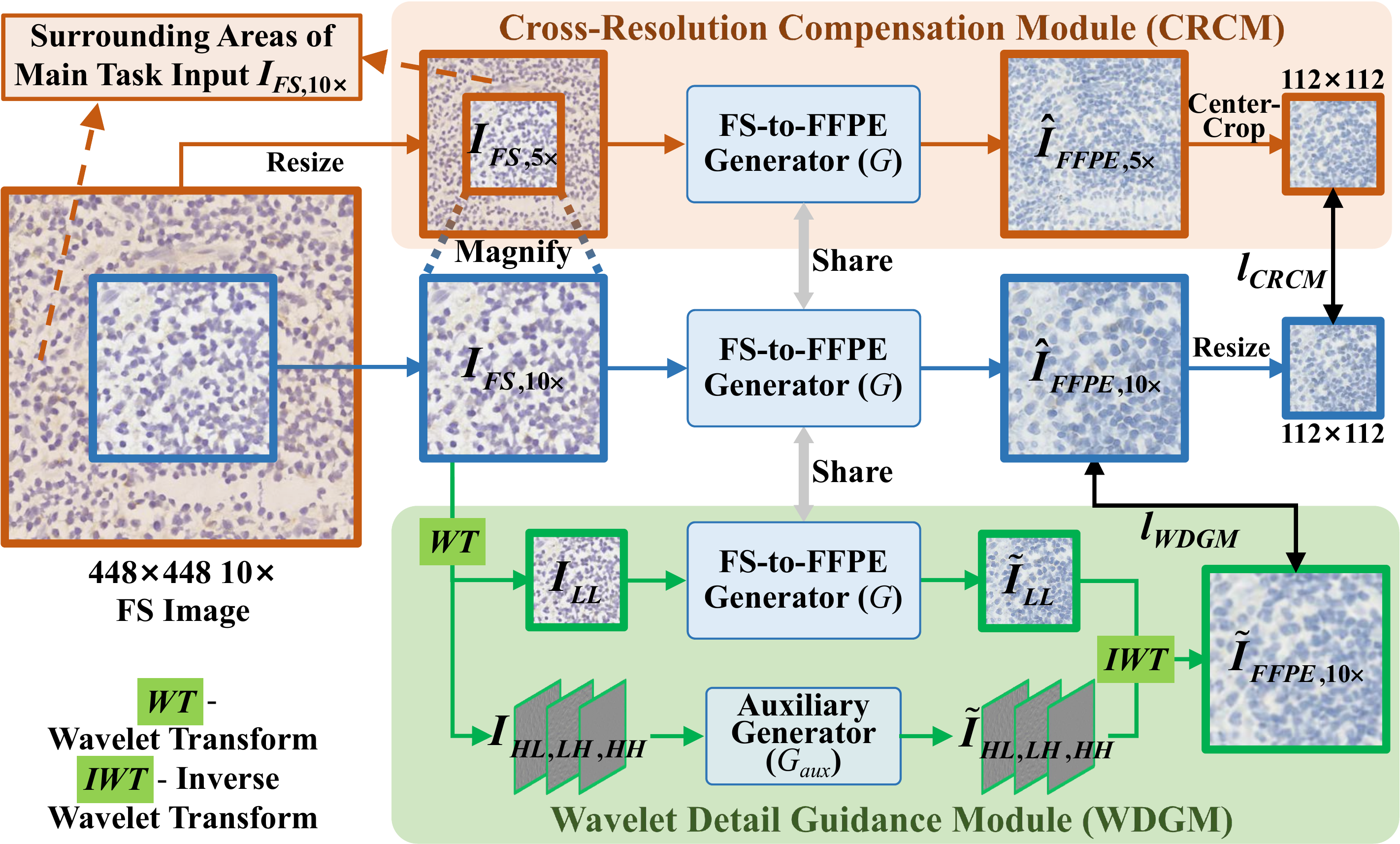}}
	\caption{\textbf{The overview of our method.} The proposed framework is composed of a main task, which can transfer the FS image to the FFPE one, as well as two auxiliary modules, CRCM in the orange box and WDGM in the green box. 
    }
	\label{overview}
\end{figure}

Since in IHC analysis, FFPE images have significantly better quality than FS images, a method of transferring FS images into FFPE ones is highly needed to improve intraoperative diagnosis based on FS images. Thus, in this paper, leveraging the high accuracy and fast speed of the GAN \cite{goodfellow2020generative}, we propose the first FS-to-FFPE stain transfer method for intraoperative IHC images, as shown in Fig. \ref{overview}. The network of our method consists of an FS-to-FFPE generator $G$, an auxiliary generator ${G_{aux}}$, and a discriminator $D$. Among these generators, $G$ is utilized to transfer the FS image into the FFPE one, and ${G_{aux}}$ is used to assist $G$ in transferring the high-frequency part of the image. The discriminator $D$ is employed to judge whether the image is real or generated by the generator. 

However, due to the inaccessibility of the paired FS and FFPE images in the clinic, existing methods, such as AI-FFPE \cite{AI-FFPE} and vFFPE \cite{vFFPE}, usually adopt the unsupervised stain transfer method. Compared with the supervised method, the accuracy of the unsupervised one may be relatively inferior, which is mainly reflected in errors of staining status (positive or negative, same as below) as well as poor details such as nuclear membrane and chromatin. To improve the accuracy, as shown in the orange and green boxes of Fig. \ref{overview}, we propose a cross-resolution compensation module (CRCM) and a wavelet detail guidance module (WDGM). Specifically, in CRCM, the network can observe more tissue surrounding the input image across multiple resolutions, so the staining status of the input image can be judged utilizing more tissue information, leading to a more accurate stain transfer. In WDGM, image details are transferred separately in the form of high-frequency information, avoiding interference with low-frequency information and significantly improving the high-frequency details in the transfer results. The detailed operations of the proposed CRCM and WDGM are illustrated as follows.


\subsection{Cross-Resolution Compensation Module (CRCM)}
Since some areas of FS images are poorly stained, their staining status may be difficult to distinguish, appearing ambiguous between positive and negative and greatly disturbing the pathologist's diagnosis. Actually, in pathological images, the staining status of an image is usually consistent with its surrounding images \cite{de2021machine}. Therefore, given that the staining status of the poorly-stained area can be judged based on its well-stained surroundings, the accuracy of the staining status judgment can be improved. However, the limitation of video memory restricts the size of the input image, making it impossible to directly feed a large surrounding area of the input image into the network. In a pathological image with a fixed size, a lower resolution provides a larger field of view, covering more tissue area than a higher resolution \cite{campanella2019clinical}. Therefore, in the proposed cross-resolution compensation module (CRCM), we reduce the resolution of the wider area surrounding the input and feed it to the network. Thus, the network can observe more tissues to judge the staining status, as shown in the orange box of Fig. \ref{overview}. 

On the whole, the main task of FS-to-FFPE stain transfer follows the blue processing flow with a resolution of $10 \times$. Additionally, FS-to-FFPE is assisted by CRCM in orange at $5 \times$ resolution. As shown on the left side of Fig. \ref{overview}, following the blue and the orange processing flows, input images of ${I_{FS,10 \times }}$ and ${I_{FS,5 \times }}$ are produced by centercropping and resizing a $10 \times $ FS image with the size of $448 \times 448$. It is noted that the tissues in ${I_{FS,5 \times }}$ contain all the tissues of ${I_{FS,10 \times }}$. In addition, ${I_{FS,5 \times }}$ also contains some surrounding tissues of ${I_{FS,10 \times }}$, which are under the orange shade. Next, ${I_{FS,10 \times }}$ and ${I_{FS,5 \times }}$ are fed to the same FS-to-FFPE generator $G$, obtaining the stain transfer results ${\hat I_{FFPE,10 \times }}$ and ${\hat I_{FFPE,5 \times }}$. The stain transfer process can be formulated as follows:
\begin{equation}
{\hat I_{FFPE,10 \times }} = G({I_{FS,10 \times }}),
\label{qianyi_high}
\end{equation}
\begin{equation}
{\hat I_{FFPE,5 \times }} = G({I_{FS,5 \times }}),
\label{qianyi_low}
\end{equation}
where $G( \cdot )$ is the FS-to-FFPE generator. 


Actually, ${\hat I_{FFPE,10 \times }}$ and the center $112 \times 112$ region of ${\hat I_{FFPE,5 \times }}$ represent the stain transfer results of the same tissue area at different resolutions. It is noted that ${\hat I_{FFPE,5 \times }}$ is obtained based on more tissue areas, including the surrounding areas of the main task. Thus, its staining status judgment can be more accurate. Therefore, we design a loss function ${l_{CRCM}}$ to make the main task result ${\hat I_{FFPE,10 \times }}$ learn from ${\hat I_{FFPE,5 \times }}$, which is as follows:
\begin{equation}
{l_{CRCM}} = {\left\| {C({{\hat I}_{FFPE,5 \times }}) - R({{\hat I}_{FFPE,10 \times }})} \right\|_1}.
\label{l_crcm}
\end{equation}
Here, ${\left\|  \cdot  \right\|_1}$ is the L1 loss, and $C( \cdot )$ is the centercrop, which crops the center $112 \times 112$ region of ${\hat I_{FFPE,5 \times }}$. In addition, $R( \cdot )$ is the resize, making the size of ${\hat I_{FFPE,10 \times }}$ to be $112 \times 112$. By adding this loss function, the stain transfer result can maintain the staining status of the input image with high fidelity. 

\subsection{Wavelet Detail Guidance Module (WDGM)}
Another problem of the existing stain transfer method is that many details, such as the nuclear membrane and nucleolus, are poorly generated. In fact, most of these details are the high-frequency information \cite{wang2022hf}, which should be transferred differently from the low-frequency information. Thus, in the proposed wavelet detail guidance module (WDGM), we transfer the high-frequency information separately, avoiding interference with low-frequency information, as shown in the green box of Fig. \ref{overview}.
Firstly, the FS image ${I_{FS,10 \times }}$ is decomposed using wavelet transform (WT), obtaining low-frequency component ${I_{LL}}$ and high-frequency components ${I_{HL}},{I_{LH}}$, and ${I_{HH}}$, which is formulated as follows:
\begin{equation}
{I_{LL}},{I_{HL}},{I_{LH}},{I_{HH}} = WT({I_{FS,10 \times }}).
\label{wt}
\end{equation}
Then, the low-frequency component and high-frequency ones are transferred separately using different generators, which are formulated as follows: 
\begin{equation}
    {{\tilde I}_{LL}} = G({I_{LL}}),
\label{qianyi_fenpin_di}
\end{equation}
\begin{equation}
    {{\tilde I}_{HL}},{{\tilde I}_{LH}},{{\tilde I}_{HH}} = {G_{aux}}({I_{HL}}),{G_{aux}}({I_{LH}}),{G_{aux}}({I_{HH}}).
\label{qianyi_fenpin_gao}
\end{equation}
Here, ${{\tilde I}_{LL}},{{\tilde I}_{HL}},{{\tilde I}_{LH}}$, and ${{\tilde I}_{HH}}$ are the transfer results of all frequency components, $G( \cdot )$ is the FS-to-FFPE generator, and ${G_{aux}}( \cdot )$ is the auxiliary generator for transferring the high-frequency components. Then, the transfer results of all components are combined using inverse wavelet transform (IWT), and the formula is as follows:
\begin{equation}
    {{\tilde I}_{FFPE,10 \times }} = IWT({{\tilde I}_{LL}},{{\tilde I}_{HL}},{{\tilde I}_{LH}},{{\tilde I}_{HH}}),
\label{iwt}
\end{equation}
where ${{\tilde I}_{FFPE,10 \times }}$ is the result of WDGM. Since high-frequency details are processed separately and interference from low-frequency information is avoided in WDGM, the details in ${{\tilde I}_{FFPE,10 \times }}$ can be better than the main task result ${{\hat I}_{FFPE,10 \times }}$ obtained by Eq. \ref{qianyi_high}. After obtaining ${{\tilde I}_{FFPE,10 \times }}$, we design a loss function ${l_{WDGM}}$, constraining the main task result ${{\hat I}_{FFPE,10 \times }}$ to be consistent with ${{\tilde I}_{FFPE,10 \times }}$. The formula of this loss function is as follows:
\begin{equation}
{l_{WDGM}} = {\left\| {{{\hat I}_{FFPE,10 \times }} - {{\tilde I}_{FFPE,10 \times }}} \right\|_1}.
\label{l_wdgm}
\end{equation}
Here, ${\left\|  \cdot  \right\|_1}$ is the L1 loss. By posing this loss function, the main task can achieve better results in the details of the stain transfer image.

\subsection{Loss Functions}
Our loss functions include the loss of the discriminator and the generator. In our discriminator, the loss function includes adversarial loss $l_{GAN}^D$ \cite{goodfellow2020generative}. 
In our generator, the loss functions include $l^G_{GAN}$ (adversarial loss \cite{goodfellow2020generative}), $l_{patchNCE}$ (a contrastive learning loss function commonly used in image style transfer \cite{CUT}), ${l_{CRCM}}$ (see Eq. \ref{l_crcm}), and ${l_{WDGM}}$ (see Eq. \ref{l_wdgm}). The total loss function of the generator is:
\begin{equation}
    {l_G} =  l_{GAN}^G + {l_{patchNCE}} + {l_{CRCM}} + {l_{WDGM}}.
	\label{LG}
\end{equation}

\section{Experiments}
\subsection{Dataset and Experimental Details}


We construct an in-house dataset using human thyroid TTF-1 (a type of IHC) slides from Peking University Shenzhen Hospital. The slides are at 10× magnification, comprising 101,856 FS and 46,623 FFPE 448×448 images. In addition, we randomly split the data into training (82,096 FS, 41,319 FFPE) and testing (19,760 FS, 5,304 FFPE) sets following the patient-level separation. Moreover, our model is implemented in Python (PyTorch) and runs on an NVIDIA RTX 4090 GPU. The training utilizes the Adam optimizer (${\beta _1} =0.5$, ${\beta _2} =0.999$), with an initial learning rate of 0.0001, which linearly decays after half of the total iterations (400K iterations). The batch size is set to 1.

\begin{figure}[t]
	\centerline{\includegraphics[width=0.85\columnwidth]{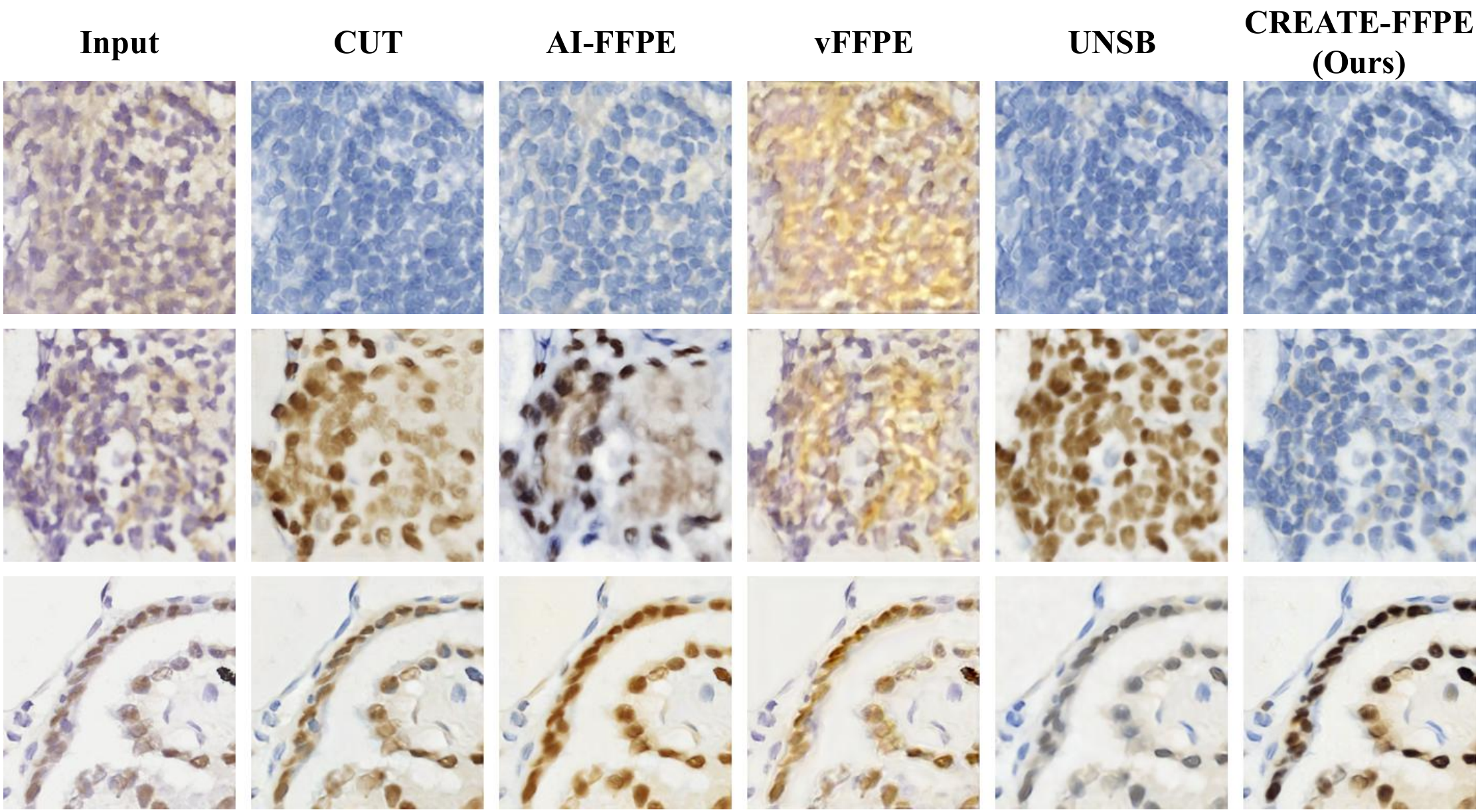}}
	\caption{\textbf{The visualization results of our method and the competing methods.} It is noted that since there are no paired FS and FFPE images in clinical practice, the ground truth is not shown here.}
	\label{Comparative_Experiments}
\end{figure}

\begin{table}[t]
\centering 
\newcolumntype{l}{>{\centering\arraybackslash}p{1.7cm}}
\caption{\textbf{The quantitative results of our CREATE-FFPE and competing methods}, including CUT, AI-FFPE, vFFPE, and UNSB.}
\begin{tabular}{c|l|l|l|l|l}
\hline
Method                                                                     & CUT              & AI-FFPE       & vFFPE   & UNSB         & \begin{tabular}[c]{@{}c@{}}CREATE-\\FFPE(Ours)\end{tabular} \\ \hline
FID (↓)                                                                       & 123.73           & 156.01        & 111.92  & {\ul 105.36} & \textbf{86.81}                                            \\ \hline
KID×100 (↓)                                                                   & 8.27             & 16.14         & 7.32    & {\ul 6.93}   & \textbf{4.65}                                             \\ \hline
\begin{tabular}[c]{@{}c@{}}Average inference\\ time per image (↓)\end{tabular} & \textbf{0.0047s} & {\ul 0.0063s} & 0.0110s & 0.0884s      & {\ul 0.0063s}                                             \\ \hline
\end{tabular}
\label{Comparative_Experiments_tab}
\end{table}

\subsection{Comparative Experiments}
In this section, we compare our method with state-of-the-art stain transfer methods for FFPE image generation, including AI-FFPE \cite{AI-FFPE} and vFFPE \cite{vFFPE}. Additionally, we include CUT \cite{CUT}, the baseline of AI-FFPE, and UNSB \cite{UNSB}, a recent high-performing image style transfer method. Moreover, following prior unpaired FS-to-FFPE stain transfer studies AI-FFPE and FastFF2FFPE \cite{AI-FFPE,FastFF2FFPE}, we use FID \cite{FID} and KID×100 \cite{KID} for quantitative evaluation.


Figure \ref{Comparative_Experiments} compares the visual results of our method with others. In the first row, vFFPE introduces yellow artifacts, obscuring nuclei and tissues. In the second row, CUT, AI-FFPE, and UNSB incorrectly stain negative nuclei (blue) as positive (brown). Additionally, in the third row, UNSB incorrectly stains positive nuclei (brown) as negative (blue-grey), potentially leading to misdiagnosis during surgery. In contrast, our method produces clear images, accurately presenting all the tissues, such as the nuclear membrane and internal texture, and maintaining the staining status (positive or negative, same as below) of the input FS image with high fidelity. Table \ref{Comparative_Experiments_tab} further validates our approach, showing a 17.6\% reduction in FID and 32.9\% reduction in KID×100 compared to the second-best method. Additionally, our CREATE-FFPE has the second shortest inference time, demonstrating its practical applicability in surgery.

\begin{figure}[t]
	\centerline{\includegraphics[width=0.8\columnwidth]{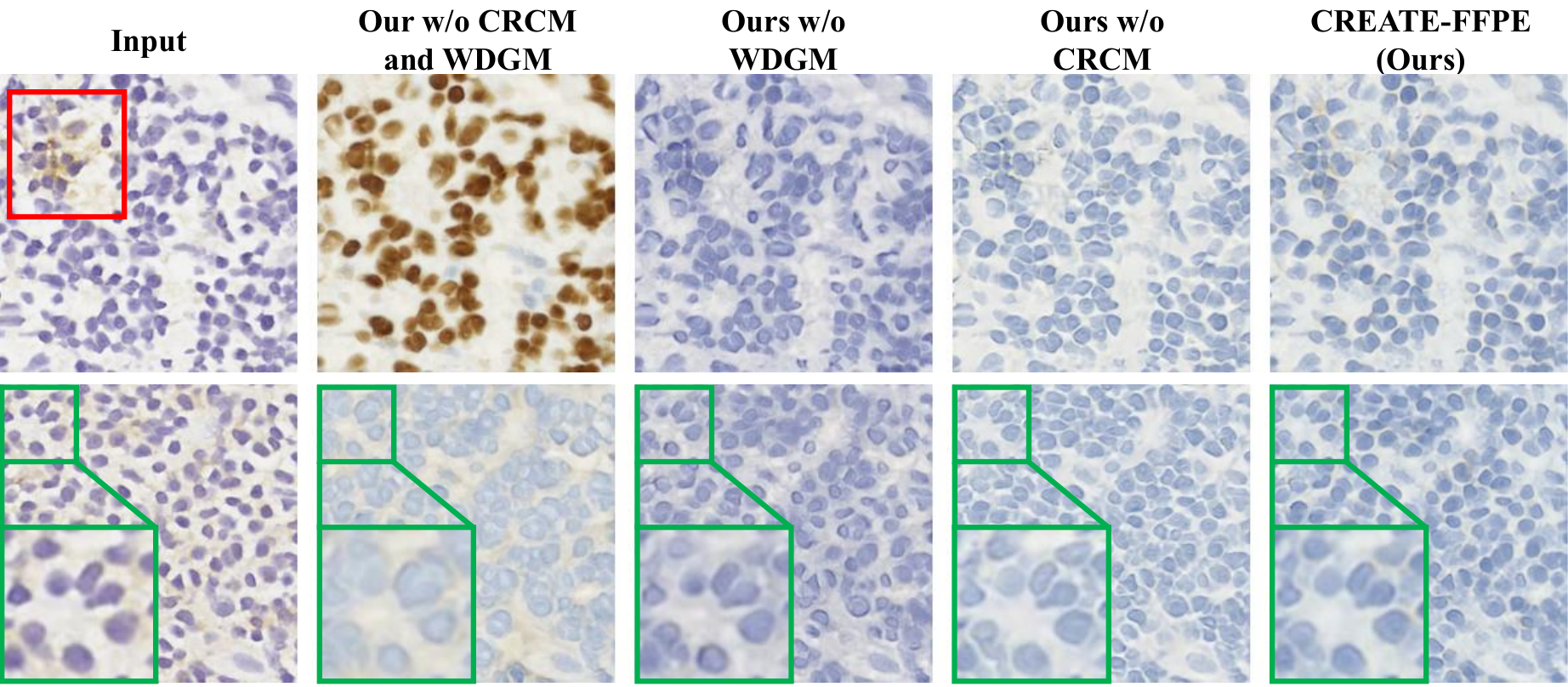}}
	\caption{\textbf{The visualization results of our ablation experiments.} To better observe the details of the nucleus, we zoom in on some areas in the second row.}
	\label{Ablation_Experiments}
\end{figure}

\begin{table}[t]
\newcolumntype{l}{>{\centering\arraybackslash}p{1.8cm}}
\centering 
\caption{\textbf{The quantitative results of our ablation experiments.} Here, we verify the effectiveness of our proposed CRCM and WDGM.}
\begin{tabular}{c|l|l|l|l}
\hline
Method  & \begin{tabular}[c]{@{}c@{}}Ours w/o \\ CRCM and \\ WDGM\end{tabular} & \begin{tabular}[c]{@{}c@{}}Ours w/o \\ WDGM\end{tabular} & \begin{tabular}[c]{@{}c@{}}Ours w/o \\ CRCM\end{tabular} & \begin{tabular}[c]{@{}c@{}}CREATE-\\FFPE(Ours)\end{tabular}           \\ \hline
FID (↓)     & 156.01                                                               & 120.23                                                   & {\ul 91.23}                                              & \textbf{86.81} \\ \hline
KID×100 (↓) & 16.14                                                                & 8.73                                                     & {\ul 5.06}                                               & \textbf{4.65}  \\ \hline
\end{tabular}
\label{Ablation_Experiments_tab}
\end{table}

\subsection{Ablation Experiments}

In our ablation experiments, we evaluate each proposed module's effectiveness. From Fig. \ref{Ablation_Experiments}, in the first row, the nuclei in the input FS image are actually negative (blue). However, this image has suffered from some brown interference, especially the area in the red box. Therefore, the baseline incorrectly identifies this image as positive and generates positive nuclei (brown), as shown in the second column. In contrast, through our proposed cross-resolution compensation module (CRCM), more well-stained tissue around the input image can be additionally employed to determine the staining status, so the output image can more accurately maintain the staining status of the input image, as shown in the third column of Fig. \ref{Ablation_Experiments}. In the second row, the baseline result suffers from unclear nuclear boundaries and textures, especially in the green zoom-in area. In contrast, our wavelet detail guidance module (WDGM) enhances high-frequency details, sharpening boundaries and revealing internal structures like nucleoli, as shown in the fourth column. Combining CRCM and WDGM ensures precise staining and fine details, reducing FID by 44.4\% and KID×100 by 71.2\%, as shown in the fifth column of Fig. \ref{Ablation_Experiments} and Tab. \ref{Ablation_Experiments_tab}.

\begin{table}[t]
\centering 
\newcolumntype{l}{>{\centering\arraybackslash}p{1.5cm}}
\caption{\textbf{The quantitative results of our downstream tasks.} In MSI prediction, We convert its FS images to FFPE ones to enhance FS quality.}
\begin{tabular}{c|c|c|c|c|c}
\hline
Method       & \begin{tabular}[c]{@{}c@{}}Top-1\\ ACC (↑)\end{tabular}           & \begin{tabular}[c]{@{}c@{}}Precision\\ (↑)\end{tabular}           & \begin{tabular}[c]{@{}c@{}}Recall\\ (↑)\end{tabular}              & \begin{tabular}[c]{@{}c@{}}F1-score\\ (↑)\end{tabular}            & \begin{tabular}[c]{@{}c@{}}AUC\\ (↑)\end{tabular}                 \\ \hline
W/o CREATE-FFPE  & \begin{tabular}[c]{@{}c@{}}0.7660\\ ±0.0019\end{tabular}          & \begin{tabular}[c]{@{}c@{}}0.6559\\ ±0.0021\end{tabular}          & \begin{tabular}[c]{@{}c@{}}0.6638\\ ±0.0127\end{tabular}          & \textbf{\begin{tabular}[c]{@{}c@{}}0.6507\\ ±0.0018\end{tabular}} & \textbf{\begin{tabular}[c]{@{}c@{}}0.7468\\ ±0.0022\end{tabular}} \\ \hline
With CREATE-FFPE & \textbf{\begin{tabular}[c]{@{}c@{}}0.7808\\ ±0.0013\end{tabular}} & \textbf{\begin{tabular}[c]{@{}c@{}}0.6824\\ ±0.0180\end{tabular}} & \textbf{\begin{tabular}[c]{@{}c@{}}0.6653\\ ±0.0036\end{tabular}} & \begin{tabular}[c]{@{}c@{}}0.6474\\ ±0.0027\end{tabular}          & \begin{tabular}[c]{@{}c@{}}0.7400\\ ±0.0076\end{tabular}          \\ \hline
\end{tabular}
\label{Downstream_Tasks_tab}
\end{table}


\subsection{Downstream Tasks}
Here, we apply our method to a microsatellite instability (MSI) prediction task. This task is a binary classification based on FS images using MobileNetV2 \cite{mobilenetv2} on a public dataset \cite{MSI-predict}. The FS images are first transferred to FFPE images, and MSI prediction is performed separately on both the original FS and transferred FFPE images. Tab. \ref{Downstream_Tasks_tab} shows significant improvements, with precision increasing by 2.7 percentage points. These results demonstrate that our CREATE-FFPE enhances FS images in downstream tasks, improving overall performance.

\section{Conclusion}

To enhance the quality of intraoperative immunohistochemistry (IHC) images obtained in the frozen-section (FS) approach, we propose CREATE-FFPE, the first framework for FS to formalin-fixed and paraffin-embedded (FFPE) stain transfer for intraoperative IHC Images. Within CREATE-FFPE, we introduce the cross-resolution compensation module (CRCM) for enriching input information and the wavelet detail guidance module (WDGM) for providing detailed guidance. In experiments, CREATE-FFPE achieves state-of-the-art performance and reduces FID by 44.4\% and KID×100 by 71.2\% by introducing CRCM and WDGM. Furthermore, our method can improve the performance of downstream microsatellite instability prediction tasks. Future work will develop a framework for FS H\&E to FFPE IHC stain transfer, allowing pathologists to determine the pathological status of tissue during surgery more cost-effectively and accurately.



%
%
%
\bibliographystyle{splncs04}

\begin{thebibliography}{10}
\providecommand{\url}[1]{\texttt{#1}}
\providecommand{\urlprefix}{URL }
\providecommand{\doi}[1]{https://doi.org/#1}

\bibitem{campanella2019clinical}
Campanella, G., Hanna, M.G., Geneslaw, L., Miraflor, A., Werneck Krauss~Silva, V., Busam, K.J., Brogi, E., Reuter, V.E., Klimstra, D.S., Fuchs, T.J.: Clinical-grade computational pathology using weakly supervised deep learning on whole slide images. Nature medicine  \textbf{25}(8),  1301--1309 (2019)

\bibitem{chao2004use}
Chao, C.: The use of frozen section and immunohistochemistry for sentinel lymph node biopsy in breast cancer. The American Surgeon  \textbf{70}(5),  414--419 (2004)

\bibitem{de2021machine}
De~Matos, J., Ataky, S.T.M., de~Souza Britto~Jr, A., Soares~de Oliveira, L.E., Lameiras~Koerich, A.: Machine learning methods for histopathological image analysis: A review. Electronics  \textbf{10}(5), ~562 (2021)

\bibitem{vFFPE}
Falahkheirkhah, K., Guo, T., Hwang, M., Tamboli, P., Wood, C.G., Karam, J.A., Sircar, K., Bhargava, R.: A generative adversarial approach to facilitate archival-quality histopathologic diagnoses from frozen tissue sections. Laboratory Investigation  \textbf{102}(5),  554--559 (2022)

\bibitem{FastFF2FFPE}
Fan, L., Sowmya, A., Meijering, E., Song, Y.: Fast ff-to-ffpe whole slide image translation via laplacian pyramid and contrastive learning. In: MICCAI. pp. 409--419. Springer (2022)

\bibitem{goodfellow2020generative}
Goodfellow, I., Pouget-Abadie, J., Mirza, M., Xu, B., Warde-Farley, D., Ozair, S., Courville, A., Bengio, Y.: Generative adversarial networks. Communications of the ACM  \textbf{63}(11),  139--144 (2020)

\bibitem{MSI-predict}
Kather, J.N., Pearson, A.T., Halama, N., J{\"a}ger, D., Krause, J., Loosen, S.H., Marx, A., Boor, P., Tacke, F., Neumann, U.P., et~al.: Deep learning can predict microsatellite instability directly from histology in gastrointestinal cancer. Nature medicine  \textbf{25}(7),  1054--1056 (2019)

\bibitem{UNSB}
Kim, B., Kwon, G., Kim, K., Ye, J.C.: Unpaired image-to-image translation via neural schrödinger bridge. In: ICLR (2024)

\bibitem{obaid2024comparison}
Obaid, Q., Nadji, M., Schlumbrecht, M., Pinto, A.: Comparison of antigenicity between frozen section vs non- frozen section tissue blocks: An immunohistochemical study of antibodies commonly used in gynecologic pathology. American journal of clinical pathology  \textbf{162}(6),  612--622 (2024)

\bibitem{AI-FFPE}
Ozyoruk, K.B., Can, S., Darbaz, B., Ba{\c{s}}ak, K., Demir, D., Gokceler, G.I., Serin, G., Hacisalihoglu, U.P., Kurtulu{\c{s}}, E., Lu, M.Y., et~al.: A deep-learning model for transforming the style of tissue images from cryosectioned to formalin-fixed and paraffin-embedded. Nature Biomedical Engineering  \textbf{6}(12),  1407--1419 (2022)

\bibitem{CUT}
Park, T., Efros, A.A., Zhang, R., Zhu, J.Y.: Contrastive learning for unpaired image-to-image translation. In: ECCV. pp. 319--345. Springer (2020)

\bibitem{mobilenetv2}
Sandler, M., Howard, A., Zhu, M., Zhmoginov, A., Chen, L.C.: Mobilenetv2: Inverted residuals and linear bottlenecks. In: CVPR. pp. 4510--4520 (2018)

\bibitem{FID}
Shmelkov, K., Schmid, C., Alahari, K.: How good is my gan? In: ECCV. pp. 213--229 (2018)

\bibitem{KID}
Sutherland, J., Arbel, M., Gretton, A.: Demystifying mmd gans. In: ICLR. pp. 1--36 (2018)

\bibitem{vahini2017intraoperative}
Vahini, G., Ramakrishna, B., Kaza, S., Murthy, R.: Intraoperative frozen section -a golden tool for diagnosis of surgical biopsies. International Clinical Pathology Journal  \textbf{4} (2017)

\bibitem{wang2022hf}
Wang, Y., Skorokhodov, I., Wonka, P.: Hf-neus: Improved surface reconstruction using high-frequency details. NeurIPS  \textbf{35},  1966--1978 (2022)

\bibitem{yigzaw2024review}
Yigzaw, D., Dagnaw, G.G.: Review of immunohistochemistry techniques: Applications, current status, and future perspectives. In: Seminars in diagnostic pathology. vol.~41, pp. 154--160. Elsevier (2024)

\end{thebibliography}

%

\end{document}